\documentclass[journal]{IEEEtran}
\ifCLASSINFOpdf
  \usepackage[pdftex]{graphicx}
  \graphicspath{{../pdf/}{../jpeg/}}
  \DeclareGraphicsExtensions{.pdf,.jpeg,.png}
\else
  \usepackage[dvips]{graphicx}
  \graphicspath{{../eps/}}
  \DeclareGraphicsExtensions{.eps}
\fi
\usepackage{amsmath}
\begin{document}
%
\title{Classification-driven Single Image Dehazing}
%
%
%

\author{{Yanting Pei, Yaping Huang, Xingyuan Zhang
        }
\thanks{Y. Pei, Y. Huang and X. Zhang are with the Beijing Key Laboratory of Traffic Data Analysis and Mining, Beijing Jiaotong University, Beijing, 100044, China.}

}

%
%

\markboth{Journal of \LaTeX\ Class Files,~Vol.~14, No.~8, August~2015}%
{Shell \MakeLowercase{\textit{et al.}}: Bare Demo of IEEEtran.cls for IEEE Journals}
%



\maketitle


\begin{abstract}
Most existing dehazing algorithms often use hand-crafted features or Convolutional Neural Networks (CNN)-based methods to generate clear images using pixel-level Mean Square Error (MSE) loss. The generated images generally have better visual appeal, but not always have better performance for high-level vision tasks, e.g. image classification. In this paper, we investigate a new point of view in addressing this problem. Instead of focusing only on achieving good quantitative performance on pixel-based metrics such as Peak Signal to Noise Ratio (PSNR), we also ensure that the dehazed image itself does not degrade the performance of the high-level vision tasks such as image classification. To this end, we present an unified CNN architecture that includes three parts: a dehazing sub-network (DNet), a classification-driven Conditional Generative Adversarial Networks sub-network (CCGAN) and a classification sub-network (CNet) related to image classification, which has better performance both on visual appeal and image classification. We conduct comprehensive experiments on two challenging benchmark datasets for fine-grained and object classification: CUB-200-2011 and Caltech-256. Experimental results demonstrate that the proposed method outperforms many recent state-of-the-art single image dehazing methods in terms of image dehazing metrics and classification accuracy. 
\end{abstract}

\section{Introduction}

Haze is a traditional atmospheric phenomenon where dust, smoke and other dry particles obscure the clarity of the atmosphere. In this age of ubiquitous smartphone usage, images captured by smartphone cameras under difficult hazy weather conditions undergo degradations that drastically affect the visual quality of images and make the images useless for sharing and usage. Meanwhile, the existence of haze dramatically degrades the visibility of outdoor images captured in the inclement weather and affects high-level computer vision tasks, such as image classifacation and other computer vision applications, such as autonomous driving, aerial photography and remote sensing.

Koschmieder et al.~\cite{koschmieder1924theorie} first propose the atmospheric scattering model, which is further developed by Narasimhan and Nayar~\cite{narasimhan2003contrast,nayar1999vision}. The atmospheric scattering model can be formally written as
\begin{eqnarray}
I(x)=J(x) \cdot t(x)+ A \cdot (1-t(x)),
\end{eqnarray}
where $x$ is the pixel coordinates, $I(x)$ is the observed hazy image, $J(x)$ is the original clear image and $A$ is the global atmospheric light. $t(x)$ is the medium transmission map and it is distance-dependent:
\begin{eqnarray}
\label{hazy-synthetic}
t(x)=e^{-\beta d(x)},
\end{eqnarray}
where $\beta$ is the atmospheric scattering coefficient and $d(x)$ is the scene depth.  The goal of image dehazing is to recover clear image $J(x)$ from hazy image $I(x)$.

Single image dehazing is an ill-posed problem and some methods try to use visual cues to capture deterministic and statistical properties of hazy images~\cite{tan2008visibility,he2011single,zhu2015fast,berman2016non,fattal2014dehazing}. Recent years, we have witnessed significant advances in image dehazing mainly due to emerging CNN-based dehazing methods~\cite{ren2016single,cai2016dehazenet,li2017aod,ren2018gated,zhang2018densely,zhang2018image}. Some works~\cite{ren2016single,cai2016dehazenet,li2017aod} remove haze based on atmospheric scattering model and some works~\cite{ren2018gated,zhang2018densely,zhang2018image} train an end-to-end model to gain a clear image.

However, those image dehazing methods above only show good visual appeal and are not necessarily useful for high-level vision tasks, such as image classification~\cite{krizhevsky2012imagenet,simonyan2014very,szegedy2015going,he2016deep,hu2017squeeze}, because they never consider information related to image classification. We find that this always leads to that the dehazed images have high performance based on dehazing evaluation metrics, such as PSNR and Structural Similarity Index (SSIM), but low performance based on the classification accuracy, or vice. Our usual purpose of image dehazing is helpful for further usage such as image classification, not just visual effects. Pei et al.~\cite{Pei2018Does} show that image dehazing achieves higher PSNR and SSIM values, but cannot improve the image classification much. Therefore, it is an important problem to develop a dehazing method that not only has better dehazing effect based on dehazing evaluation metrics (e.g. PSNR and SSIM) and but also has higher classification performance.

A common approach in computer vision is to separate low-level vision tasks (e.g. image dehazing) from high-level vision tasks (e.g. image classification) and solve them independently. In this paper, we propose an unified method considering both image dehazing and classification tasks. We jointly minimize the image dehazing loss and the classification loss. With the guidance of image classification, the dehazing network is able to further improve visual quality and generate more visually appealing outputs and have better classification accuracy, which demonstrates the importance of high-level information for image dehazing. We achieve this by enforcing the dehazing sub-network to adaptively learn those features which can lead to improved visual appeal and image classification performance.
 
The main contributions of this paper are as follows:
\begin{itemize}
\item We first propose an end-to-end unified CNN architecture combining dehazing and classification for image dehazing and the CNN architecture can be optimized jointly. 

\item Instead of using general CGAN, we use a classification-driven CGAN sub-network and a classification sub-network for improve the dehazing and classification performance of the dehazed images simultaneously.

\item We conduct extensive experiments on synthesized hazy images, which show that our method achieves best performance both on images dehazing metrics (PSNR and SSIM) and classification accuracies of AlexNet, VGGNet and ResNet. Besides, we test our model on real hazy images and it has good visual appeal, which indicates that the effectiveness of our proposed model.

\end{itemize} 

\section{Related Work} \label{sec:related}

In this section, we will briefly review the most related works: image dehazing, image classification and generative adversarial network.

\subsection{Image Dehazing}
 
Single image dehazing is an extremely ill-posed and challenging problem. Single image haze removal has made significant progresses recently, due to the use of better assumptions and priors~\cite{tan2008visibility,he2011single,fattal2014dehazing,zhu2015fast,berman2016non}. Specifically, Tan et al.~\cite{tan2008visibility} propose a local contrast maximizing method based on markov random field for haze removal under the assumption that the local contrast of the haze-free image is much higher than that of hazy image. Although contrast maximizing approach is able to achieve impressive results, it tends to produce over-saturated images. Inspired by dark-object subtraction technique, He et al.~\cite{he2011single} propose a dehazing method based on dark channel prior that is at least one color channel has some pixels with very low intensities in most of non-haze patches. Meng et al.~\cite{meng2013efficient} propose an effective regularization dehazing method to restore the haze-free image by exploring the inherent boundary constraint. Tang et al.~\cite{tang2014investigating} combine four types of haze relevant features with random forests to estimate the transmission. The four types of haze relevant features are dark channel, local max contrast, hue disparity and local max saturation. Fattal~\cite{fattal2014dehazing} proposes a dehazing method relying on a generic regularity in natural images in which pixels of small image patches exhibit one-dimensional distributions in RGB space, known as color-lines. Zhu et al.~\cite{zhu2015fast} present a single image haze removal algorithm using the color attenuation prior by creating a linear model for modeling the scene depth of the hazy image under this prior. Berman et al.~\cite{berman2016non} introduce a haze removal method based on a non-local prior, by assuming that colors of a haze-free image are well approximated by a few hundred of distinct colors in the form of tight clusters in RGB space. In a hazy image, these tight color clusters change due to haze and form lines in RGB space that pass through the airlight coordinate.

CNNs have witnessed prevailing success in computer vision tasks and are recently introduced to haze removal~\cite{ren2016single,cai2016dehazenet,li2017aod,ren2018gated,zhang2018densely,zhang2018image}. Ren et al.~\cite{ren2016single} propose a multi-scale deep neural network for haze removal, and the network consists of a coarse-scale sub-network for a holistic transmission map and a fine-scale sub-network for local refinement. Cai et al.~\cite{cai2016dehazenet} adopt CNN-based deep architecture, whose layers are specially designed to embody the established priors in image dehazing and it is constructed by three convolution layers, a max-pooling, a Maxout unit and a BReLU activation function. Li et al.~\cite{li2017aod} propose a light-weight CNN designation based on a re-formulated atmospheric scattering model. Instead of estimating the transmission matrix and the atmospheric light separately as most previous models did, Ren et al.~\cite{ren2018gated} propose an end-to-end trainable neural network that consists of an encoder and a decoder. The encoder is exploited to capture the context of the derived input images that are White Balance, Contrast Enhancing, and Gamma Correction, while the decoder is employed to estimate the contribution of each input to the final dehazed result. Zhang and Patel~\cite{zhang2018densely} directly embed the atmospheric scattering model into the network and propose a new edge-preserving densely connected encoder-decoder structure with multi-level pyramid pooling module for estimating the transmission map and this network is optimized using a newly introduced edge-preserving loss function. Zhang ~\cite{zhang2018image} proposes a dehazing method based on a conditional generative adversarial network, where the clear image is estimated by an end-to-end trainable neural network.

\subsection{Image Classification}


In recent years, image classification has made significant progress, partly due to the creation of large-scale hand-labeled datasets such as ImageNet~\cite{deng2009imagenet}, and the development of deep convolutional neural networks~\cite{krizhevsky2012imagenet}. Current state-of-the-art image classification methods focus on training feed forward convolutional neural networks using ``very deep'' structure~\cite{simonyan2014very,szegedy2015going,he2016deep}. VGGNet~\cite{simonyan2014very}, Inception~\cite{szegedy2015going} and residual learning~\cite{he2016deep} have been proposed to train very deep neural networks, resulting in excellent image-classification performances on clear natural images. Liu et al.~\cite{liu2015treasure} propose a cross-convolutional-layer pooling method for image classification. Wang et al.~\cite{wang2016cnn} combine CNN with recurrent neural networks (RNN) for improving the image classification performance. Durand et al.~\cite{durand2017wildcat} study three important visual recognition tasks, image classification, weakly supervised point-wise object localization and semantic segmentation in an integrative way. Wang et al.~\cite{wang2017residual} develop a convolutional neural network using attention mechanism for image classification. Hu et al.~\cite{hu2017squeeze} propose an architectural unit based on the channel relationship, which adaptively recalibrates the channel-wise feature responses by explicitly modeling interdependencies between channels.

\subsection{Generative Adversarial Network}

Generative Adversarial Networks (GANs) have become more and more popular recently.
Goodfellow et al.~\cite{Goodfellow2014Generative} first propose GAN~\cite{Goodfellow2014Generative} to synthesize realistic images by learning the distribution of training images. Initially, the training of GANs is unstable, which often results in artifacts in the synthesized images. Incorporating conditional information in GAN results in more effective learning~\cite{Sohn2015Learning}. The conditioning variables augmenting information increases the stability of learning process and improves the representation capability of the generator. Different from original GAN~\cite{Goodfellow2014Generative}, the CGAN algorithm learns to generate a clear image $J$ from an input image $I$ and random noise $z$ by optimizing the objective function. The CGAN has been made great progress in image processing field such as super-resolution~\cite{Ledig2016Photo}, image inpainting~\cite{Xu2017Learning} and style transfer~\cite{Johnson2016Perceptual}. Raymond et al.~\cite{Yeh2016Semantic} propose a semantic image inpainting algorithm using a CGAN. In image super-resolution, Ledig et al.~\cite{Ledig2016Photo} modify the GAN formulation by introducing pixel-wise content loss and perceptual loss~\cite{Johnson2016Perceptual} to generate high quality images. Zhang et al.~\cite{zhang2017image} use the pixel-wise content loss and perceptual loss in CGAN to solve image deraining problem. Based on CGAN, Zhang~\cite{zhang2018image} also proposes an architecture for image dehazing.
  
\section{Our method}

Instead of directly learning a mapping from an input hazy image to a dehazed image by using MSE loss, which can generate dehazed images that always have better performance in terms of PSNR and SSIM metrics, we aim to generate dehazed images that have better performance both on dehazing metrics and image classification accuracy. To this end, we introduce the classification-driven CGAN sub-network and the classification sub-network. The proposed network is composed of three important parts: dehazing sub-network, classification-driven CGAN sub-network and the classification sub-network, which serves as distinct purposes. In this section, we first introduce the architecture of the proposed network. Then we describe each part in detail as well as the loss function.

\subsection{Overview}

\begin{figure*}[htbp]
\centering
\includegraphics[height=8.5cm]{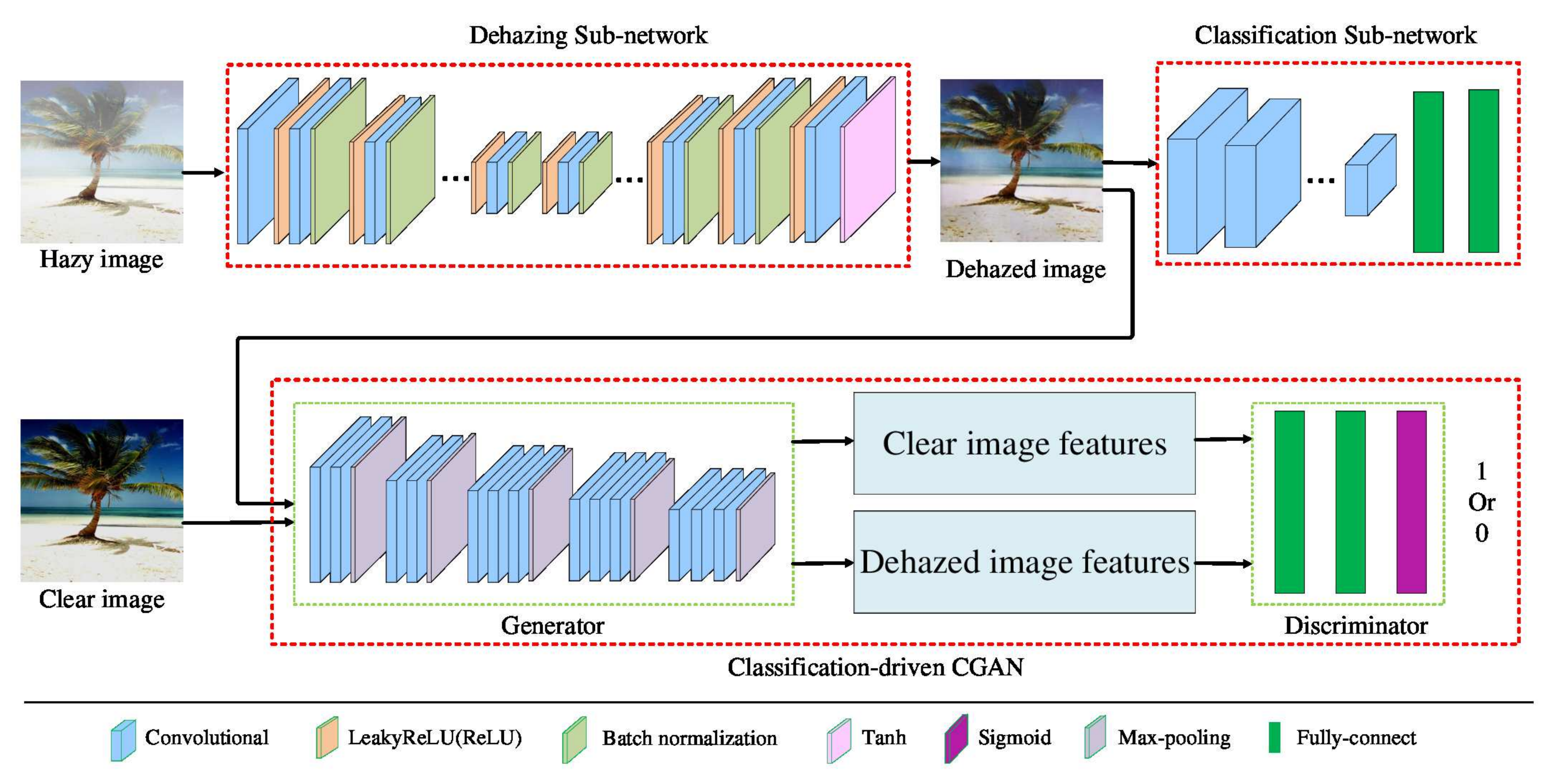}
\vspace{-1em}
\caption{The overview of our network architecture, which is composed of three parts: image dehazing sub-network with MSE loss, image classification-driven CGAN sub-network with GAN loss and image classification sub-network with Cross Entropy loss.}
\label{Fig:overview}
\end{figure*}

We propose an unified network that can be used not only to image dehazing but also to image classification, which takes a hazy image as input and can output the dehazed image as well as the image category. The proposed network is composed of three parts: image dehazing sub-network (DNet), image classification-driven CGAN sub-network (CCGAN) and image classification sub-network (CNet). The overview of our method is shown in Fig.~\ref{Fig:overview}. For the DNet, we use the commonly used MSE loss to generate dehazed image that aims to have visual appeal. For CCGAN, we use the GAN loss to generate dehazed image that aims to have better classification performance. For CNet, we use Cross Entropy loss to generate dehazed image that aims to further improve the classification performance.

\subsection{Dehazing Sub-network}
\label{dehazing sub-network}
 
The purpose of the dehazing sub-network is to generate a clear image from an input hazy image. Therefore, it should not only preserve the structure and detail information of an input image but also remove the haze as much as possible. Motivated by “ResNet” [14] and “U-Net” [25], we introduce skip connections of the symmetric layers to break through the bottleneck of information in decoding process. The details of the generator structures and parameter settings are shown in Table~\ref{table:dehazing sub-network}. Each layer of the encoding process consists of the convolution, batch normalization and LeakyReLU. Each layer of the decoding process is composed of deconvolution, batch normalization and ReLU. The size of the input and output in the generator is set to be $256\times256\times3$. The size of the input in the discriminator is set to be $256\times256 \times6$ and the size of its output is $256\times256\times1$.
 

\subsection{Classification-driven CGAN Sub-network}
\label{classification-driven GAN sub-network}

In order to make the generated image have better classification performance, we introduce the classification-driven CGAN sub-network. For learning a good generator G so as to fool the learned discriminator D and make the discriminator D good enough to distinguish the real and the fake, the proposed method alternatively updates G and D. Given an input hazy image $I$ and a random noise vector $z$, conditional GAN aims to learn a mapping function to generate dehazed image $J^*$ by solving the following optimization problem:

\begin{eqnarray*}
\mathop{min}\limits_{G}\mathop{max}\limits_{D}=E_{I\sim p_{data(I)},z\sim p_{(z)}}[log(1-D(I,G(I,z)))]\\
+E_{I\sim p_{data(I,J)}}[log(D(I,J^*))]
\end{eqnarray*}

\subsubsection{Generator}

Instead of generating good dehazed image as common CGAN, the function of the generator in this paper is to generate good features of an image. As shown in Fig.~\ref{Fig:overview}, we feed the clear image and the dehazed image to the generator and gain the features of those two images, respectively. Then, we use discriminator to discriminate which features come from the clear image and which features come from the dehazed image. The network structure of generator uses VGGNet that removes the fully connected layers. Due to the size of the dehazed image is $256 \time 256$, The size of the features in the last layer is $8 \times 8$, instead of $7 \times 7$. Note that we can also use other network structure.
 
\subsubsection{Discriminator}

The discriminator is used to distinguish whether the features come from a clear image (real) or a dehazed image (fake). Therefore, we develop a two fully connected layers network. For the final layer of the discriminator, we apply a sigmoid function to the feature maps so that the probability score can be normalized into [0,1].

\subsection{Classification Sub-network}
\label{classification sub-network}

In order to further improve the classification performance, we introduce a classification sub-network. We jointly train dehazing sub-networks, classification-driven CGAN sub-network and classification sub-networks to achieve better performance not only for PSNR and SSIM, but also for classification performance. The predicted output image (dehazed image) from dehazing sub-network is fed as an input to the classification sub-network. The classification sub-network can help the dehazing sub-network to generate clearer dehazed image that has better classification performance.

\begin{table*}[htbp]
\renewcommand{\arraystretch}{1.1}
\caption{The details of the generator structure and parameter.}
\label{table:dehazing sub-network}
\centering
\small
\setlength{\tabcolsep}{1.0mm}{
\begin{tabular}{|l|l|l|l|l|l|l|l|l|l|l|l|l|l|l|l|l|}
\hline
Layer &Conv &Conv &Conv &Conv &Conv &Conv &Conv &Conv &Dconv &Dconv &Dconv &Dconv &Dconv &Dconv &Dconv &Dconv\\
\hline
Kernel size &4 $\times$ 4 &4 $\times$ 4 &4 $\times$ 4 &4 $\times$ 4 &4 $\times$ 4 &4 $\times$ 4 &4 $\times$ 4 &4 $\times$ 4 &4 $\times$ 4 &4 $\times$ 4 &4 $\times$ 4 &4 $\times$ 4 &4 $\times$ 4 &4 $\times$ 4 &4 $\times$ 4 &4 $\times$ 4\\
\hline
Stride &2 $\times$ 2 &2 $\times$ 2 &2 $\times$ 2 &2 $\times$ 2 &2 $\times$ 2 &2 $\times$ 2 &2 $\times$ 2 &2 $\times$ 2 &2 $\times$ 2 &2 $\times$ 2 &2 $\times$ 2 &2 $\times$ 2 &2 $\times$ 2 &2 $\times$ 2 &2 $\times$ 2 &2 $\times$ 2 \\
\hline
Padding &1 $\times$ 1 &1 $\times$1 &1 $\times$ 1 &1 $\times$ 1 &1 $\times$1 &1 $\times$ 1 &1 $\times$ 1 &1 $\times$ 1 &1 $\times$1 &1 $\times$ 1 &1 $\times$ 1 &1 $\times$ 1 &1 $\times$ 1 &1 $\times$ 1 &1 $\times$ 1 &1 $\times$ 1 \\
\hline
Channel &64 &128 &256 &512 &512 &512 &512 &512 &512 &512 &512 &512 &256 &128 &64 &3\\
\hline
\end{tabular}
}
\end{table*}

\subsection{Loss Function}

Let $I(x)$ and $J(x)$ denote the hazy images and the corresponding clear images. A straightforward way to train the dehazing network is to directly utilize the MSE loss $L_{MSE}$, which is given by:
\begin{eqnarray*}
L_{MSE}=\frac{1}{S}\sum_{x=1}^{N}(J(x)-J^*(x))^2
\end{eqnarray*}
where $J^*(x)$ is the dehazed image and $S$ is the size of $I(x)$. However, we find that the method using this function is not able to make the dehazed image have better performance both on PSNR, SSIM and classification accuracy. 

In order to recover realistic images, we introduce the CCGAN, the loss of which is given by:
\begin{eqnarray*}
L_{GAN}=\frac{1}{S}\sum_{x=1}^{N}log(1-D(I(x),J^*(x)))
\end{eqnarray*}

Besides, in order to improve the image classification performance, we introduce the Cross Entropy loss $L_{CE}$. Where $a$ is the output of the last fully-connected layer of CNet that is fed to a $C$-way softmax function and $C$ is the number of classes.
\begin{eqnarray*}
L_{CE}=-\sum_{i=1}^{C}y_ilog(P_i)
\end{eqnarray*}

\begin{eqnarray*}
P_i=\frac{exp(a_i)}{\sum_{r=1}^{C}exp(a_r)}
\end{eqnarray*}

Finally, we combine the MSE loss, the GAN loss and the Cross Entropy loss to regularize the proposed network, which is defined as

\begin{eqnarray*}
L=a*L_{MSE}+b*L_{GAN}+c*L_{CE}
\end{eqnarray*}

We learn all parameters of the network jointly in an end-to-end fashion.

\section{Experiments}

In this section, we first introduce datesets, experimental details and evaluation metrics briefly. Then we quantitatively and qualitatively evaluate our method against several state-of-the-art algorithms on synthetic and real-world hazy images.

\subsection{Datasets}

In this section, we evaluate various image dehazing methods on the hazy images synthesized from CUB-200-2011~\cite{Wah2011The} dataset and on the hazy images synthesized from Caltech-256~\cite{griffin2007caltech} dataset, which have been widely used for evaluating image classification algorithms. We synthesize hazy images following~\cite{cai2016dehazenet}. 

CUB-200-2011 dataset contains 11,788 images from 200 classes, which has 5994 training images and 5794 testing images. Among the training images, 20\% images are used as a validation set. Caltech-256 dataset contains 30,607 images from 257 classes. In Caltech-256, we select 60 images from each class as training images, and the rest as test images. Among the training images, 20\% per class are used as a validation set. We follow this to split the synthetic hazy image data: an image is in training set if it is synthesized from an image in the training set and in testing set otherwise. 

\subsection{Experimental Details and Evaluation Metrics}
 
In training process, we empirically set $a=500$, $b=1$ and $c=1$. The learning rate is set to be 0.0002. We use the Adam optimization method~\cite{kingma:adam} to train our network. While the proposed CNet can use ResNet-50, ResNet-101, VGGNet or other models, for convenience, we use ResNet-50 in this paper. We set the parameter $\beta = 2$ in Eq.~\ref{hazy-synthetic}.
 
We will quantitatively evaluate our dehazing method on the synthetic datasets and compare it with several state-of-the-art single image dehazing methods not only using PSNR and SSIM which are widely used for evaluating the performance of image dehazing when the ground-truth haze-free image is available, but also using classification accuracy of AlexNet~\cite{krizhevsky2012imagenet}, VGGNet-16~\cite{szegedy2015going} and ResNet-50~\cite{he2016deep}. The AlexNet, VGGNet-16 and ResNet-50 architectures are pre-trained on ImageNet dataset that consists of 1,000 classes with 1.2 million training images. For fair and comprehensive comparison, we have two strategies. First, we fine-tune AlexNet, VGGNet-16 and ResNet-50 on original clear images in CUB-200-2011 and Caltech-256 datasets, respectively. Note that we change the number of channels in the last fully connected layer from 1,000 to $N$, where $N$ is the number of classes in our datasets. We use the fine-tuned model as a classifier to test the dehazed images of our method and other state-of-the-art methods. Second, we use the CNet in our network structure as a classifier to classify the dehazed images of our method and the state-of-the-art dehazing methods.

\subsection{Quantitative and Qualitative Comparison on Synthetic Hazy images}

We compare our proposed method with nine state-of-the-art dehazing methods: Dark Channel Prior (\textbf{DCP})~\cite{he2011single}, Boundary Constrained Context Regularization (\textbf{BCCR})~\cite{meng2013efficient}, Color Attenuation Prior (\textbf{CAP})~\cite{zhu2015fast}, Non-local Image Dehazing (\textbf{NLD})~\cite{berman2016non}, \textbf{DehazeNet}~\cite{cai2016dehazenet}, Multi-Scale Convolutional Neural Networks (\textbf{MSCNN})~\cite{ren2016single}, All-in-One Dehazing Network (\textbf{AOD})~\cite{li2017aod}, Gated Fusion Network (\textbf{GFN})~\cite{ren2018gated} and Single Image Dehazing via Conditional Generative Adversarial Network (\textbf{ID-CGAN})~\cite{zhang2018image}. We compare the performance of different methods on the test images from the synthetic datasets quantitatively and qualitatively. As the ground truth is available for these test hazy images, we can calculate the quantitative measures such as PSNR and SSIM. Besides, in order to test whether the image classification performance is improved or not for the dehazed images, we also calculate the classification accuracies (\%) of AlexNet, VGGNet-16 and ResNet-50. The quantitative results are shown in Table.~\ref{table:CUB-200-2011} and Table.~\ref{table:Caltech-256}. It can be clearly observed that the proposed method is able to achieve superior quantitative performance. Our proposed network structure is not only used for image dehazing, but also used for image classification. We use our CNet as a classifier to test the dehazed images of our dehazing method and other state-of-the-art dehaizng methods, the results are shown in the last column in Table.~\ref{table:CUB-200-2011} and Table.~\ref{table:Caltech-256}. We can see that our CNet can improve the classification performance significantly, especially for fine-grained image classification shown in the last column in Table.~\ref{table:CUB-200-2011}. Experiments show that our method is very useful both for image dehazing and classification.

\begin{table}[htbp]
\renewcommand{\arraystretch}{1.2}
\caption{The dehazing and classification results of state-of-the-art and our proposed methods on CUB-200-2011 dataset.}
\label{table:CUB-200-2011}
\centering
\small
\setlength{\tabcolsep}{0.8mm}{
\begin{tabular}{|l|c|c|c|c|c|c|}
\hline
 &PSNR &SSIM & AlexNet &VGGNet &ResNet &CNet\\
\hline
DCP &16.3789 &0.7727 &35.6 &55.5 &54.8 &61.2\\
\hline
BCCR &16.2971 &0.7380 &35.9 &56.7 &56.9 &62.6\\
\hline
CAP &14.7763 &0.7581 &29.9 &57.1 &55.5 &61.6\\
\hline
NLD &14.7999 &0.6882 &32.7 &56.5 &55.9 &62.4\\
\hline
DehazeNet &15.2055 &0.7735 &29.8 &58.3 &57.9 &63.2\\
\hline
MSCNN &15.4825 &0.7573 &35.1 &57.5 &58.3 &64.2\\
\hline
AOD &13.9105 &0.7570 &24.0 &55.1 &54.0 &59.3\\
\hline
GFN &15.0244 &0.7764 &36.5 &57.5 &56.8 &63.8\\
\hline
ID-CGAN &17.2472 &0.7710 &36.9 &58.8 &59.3 &64.5\\
\hline
Ours &\textbf{21.2995} &\textbf{0.8541} &\textbf{41.0} &\textbf{60.6} &\textbf{59.7} &\textbf{67.7}\\
\hline
\end{tabular}
}
\end{table}

\begin{table}[htbp]
\renewcommand{\arraystretch}{1.2}
\caption{The dehazing and classification results of state-of-the-art and our proposed methods on Caltech-256 dataset.}
\label{table:Caltech-256}
\centering
\small
\setlength{\tabcolsep}{0.8mm}{
\begin{tabular}{|l|c|c|c|c|c|c|}
\hline
 &PSNR &SSIM & AlexNet &VGGNet &ResNet &CNet\\
\hline
DCP &17.5894 &0.7810 &56.6 &73.8 &78.9 &80.7\\
\hline
BCCR &15.7325 &0.7221 &55.4 &71.9 &77.4 &79.4\\
\hline
CAP &15.8546 &0.7718 &53.9 &74.8 &80.5 &81.3\\
\hline
NLD &16.5254 &0.7413 &56.1 &73.9 &79.3 &80.6\\
\hline
DehazeNet &15.9763 &0.7805 &54.8 &75.0 &81.0 &81.9\\
\hline
MSCNN &16.0469 &0.7640 &56.3 &74.6 &79.8 &80.9\\
\hline
AOD &14.6293 &0.7596 &49.8 &73.4 &79.8 &80.5\\
\hline
GFN &16.9605 &0.7951 &57.5 &74.5 &80.3 &81.5\\
\hline
ID-CGAN &16.7434 &0.7622 &58.0 &74.7 &78.8 &80.9\\
\hline
Ours &\textbf{21.7074} &\textbf{0.8477} &\textbf{59.3} &\textbf{76.2} &\textbf{81.1} &\textbf{82.6}\\
\hline
\end{tabular}
}
\end{table}

\begin{figure*}[htbp]
\centering
\includegraphics[height=9.3cm]{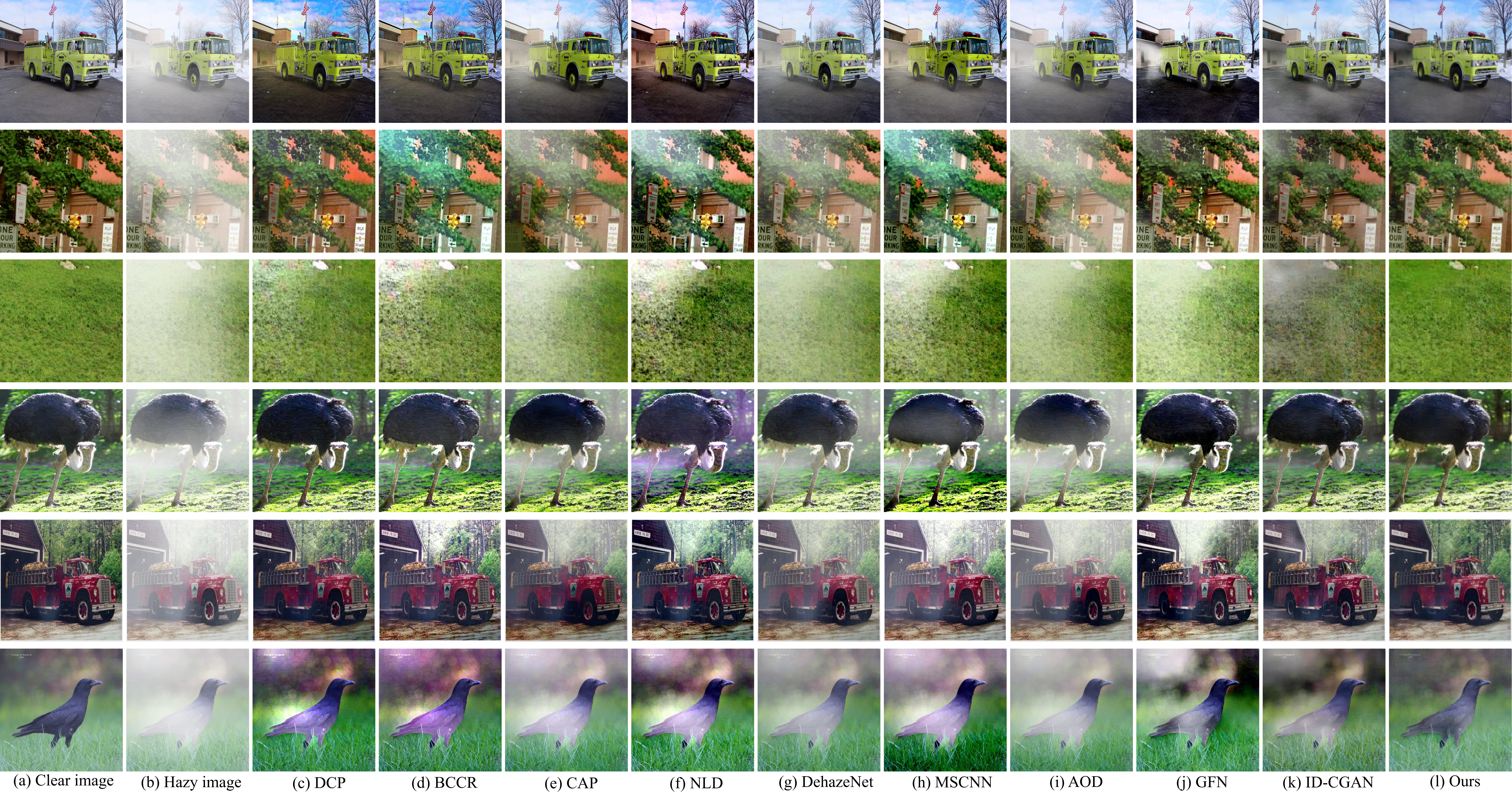}
\vspace{-1em}
\caption{Qualitative results of synthetic hazy images using several state-of-the-art dehazing methods and our proposed method.}
\label{figure:qualitative-comparison}
\end{figure*}

\begin{figure}[!b]
\centering
\includegraphics[height=3.1cm]{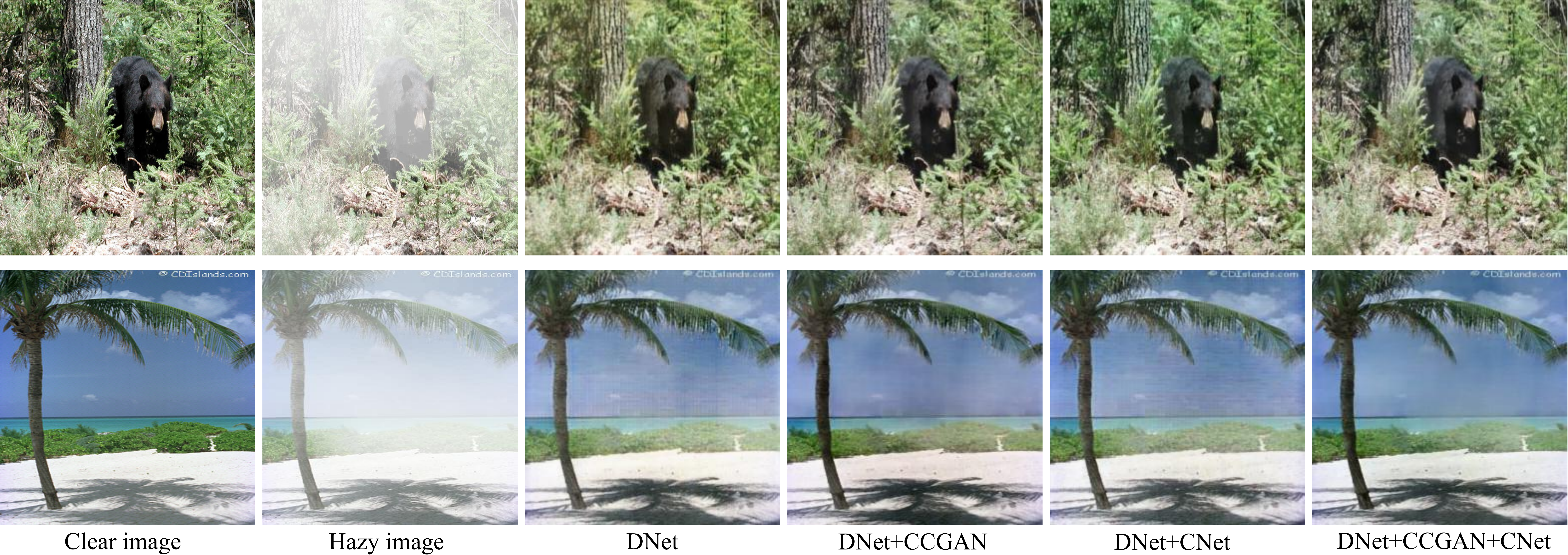}
\caption{The effect of the proposed network with different parts.}
\label{figure:different_parts}
\end{figure}

\begin{figure*}[htbp]
\centering
\includegraphics[height=5.2cm]{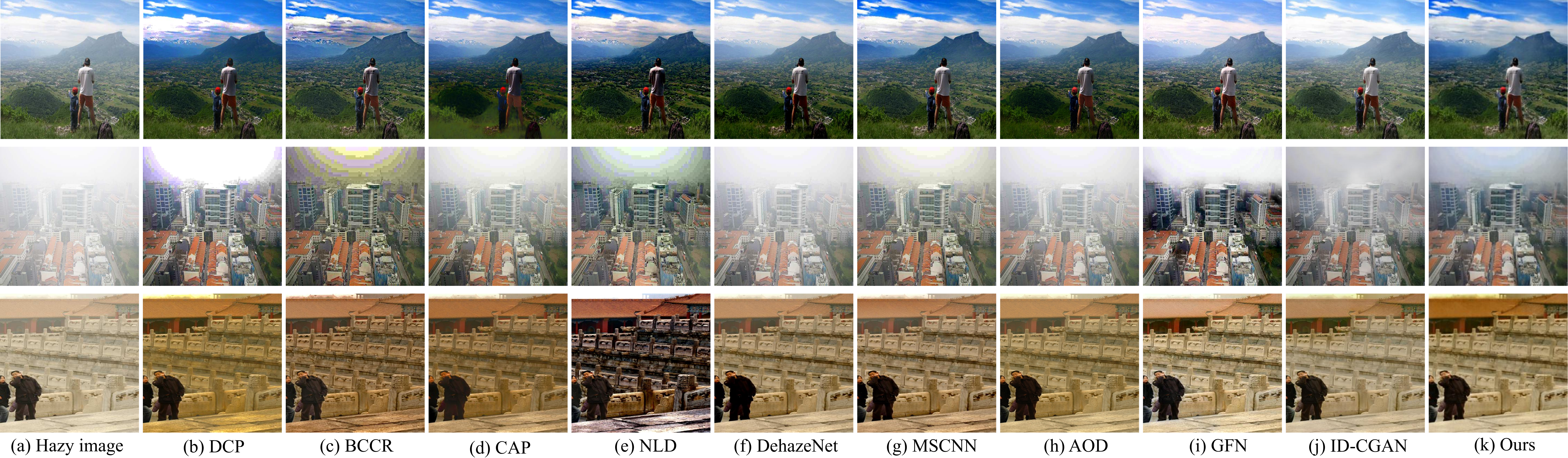}
\vspace{-1em}
\caption{Qualitative results of real hazy images using several state-of-the-art dehazing methods and our proposed method.}
\label{figure:qualitative-realhaze}
\end{figure*}

To visually demonstrate the improvements obtained by the proposed method on the synthetic dataset, we sample some dehazing results, as shown in~Fig.~\ref{figure:qualitative-comparison}. While DCP~\cite{he2011single}, BCCR~\cite{meng2013efficient}, CAP~\cite{zhu2015fast} and NLD~\cite{berman2016non} are able to remove the haze, they remove haze excessively (e.g., the dehazed images in the first row and in the third and sixth columns and the dehazed images in the fifth row and in the third column) and make the image have color distortion (e.g., the dehazed image in the forth row and in the sixth column and the dehazed images in the sixth row and in the third to sixth columns). The CNN-based methods are able to either reduce the intensity of haze or remove the haze in parts, but they fail to completely remove the haze. GFN~\cite{ren2018gated} removes haze excessively (e.g., the first row and the tenth column) and ID-CGAN~\cite{zhang2018image} dehazing method leads to color distortion (e.g., the third row and the eleventh column). In contrast to the other methods, our proposed method is able to successfully remove majority of the haze while guarantees no color distortion and the dehazed images using our method are closest to the ground truth images, as shown in the last column in~Fig.~\ref{figure:qualitative-comparison}.

\subsection{Ablation Study}

To better demonstrate the effectiveness of each part of our proposed method, we implement detailed ablation experiments by considering the combination of three factors: dehazing sub-network, classification-driven CGAN sub-network and the classification sub-network. The results are shown in Table~\ref{table:CUB-200-2011-parts} and Table~\ref{table:Caltech-256-parts}. \emph{\textbf{DNet}} refers to use dehazing sub-network only, \emph{\textbf{DNet+CCGAN}} refers to use dehazing sub-network and classfication-driven CGAN, \emph{\textbf{DNet+CNet}} refers to use dehazing sub-network and classification sub-network, and \emph{\textbf{DNet+CCGAN+CNet}} refers to use all parts. 

We can see that \emph{\textbf{DNet+CCGAN+CNet}} achieves the best performance of image dehazing both in PSNR and SSIM and classification accuracy. Compared with \emph{\textbf{DNet}}, when we add the classification sub-network (\emph{\textbf{DNet+CNet}}) and the classification-driven GAN (\emph{\textbf{DNet+CCGAN}}) respectively, not only the dehazing performance is improved, but also the classification accuracies are improved. These ablation study demonstrates that the classification-driven CGAN sub-network and the classification sub-network are effective for image dehazing.

\begin{table}[htbp]
\renewcommand{\arraystretch}{1.2}
\caption{The dehazing and classification results of the proposed network with different parts on CUB-200-2011 dataset.}
\label{table:CUB-200-2011-parts}
\centering
\scriptsize
\setlength{\tabcolsep}{1.1mm}{
\begin{tabular}{|l|c|c|c|c|c|c|}
\hline
 &PSNR &SSIM & AlexNet &VGGNet &ResNet &CNet\\
\hline
DNet &20.2032 &0.7740 &39.3 &56.6 &54.4 &63.2\\
\hline
DNet+CCGAN &21.1763 &0.8531 &40.2 &59.1 &58.0 &66.3\\
\hline
DNet+CNet &21.0780 &0.8459 &40.9 &58.0 &55.4 &64.7\\
\hline
DNet+CCGAN+CNet &\textbf{21.2995} &\textbf{0.8541} &\textbf{41.0} &\textbf{60.6} &\textbf{59.7} &\textbf{67.7}\\
\hline
\end{tabular}
}
\end{table}

\begin{table}[htbp]
\renewcommand{\arraystretch}{1.2}
\caption{The dehazing and classification results of the proposed network with different parts on Caltech-256 dataset.}
\label{table:Caltech-256-parts}
\centering
\scriptsize
\setlength{\tabcolsep}{1.1mm}{
\begin{tabular}{|l|c|c|c|c|c|c|}
\hline
 &PSNR &SSIM & AlexNet &VGGNet &ResNet &CNet\\
\hline
DNet &21.2674 &0.8194 &59.3 &74.7 &80.2 &81.9\\
\hline
DNet+CCGAN &21.7035 &0.8466 &59.2 &75.4 &81.1 &82.5\\
\hline
DNet+CNet &21.4832 &0.8444 &\textbf{59.4} &75.5 &81.0 &82.6\\
\hline
DNet+CCGAN+CNet &\textbf{21.7074} &\textbf{0.8477} &59.3 &\textbf{76.2} &\textbf{81.1} &\textbf{82.6}\\
\hline
\end{tabular}
}
\end{table}

Fig.~\ref{figure:different_parts} shows some dehazed images with different parts. We can see that when only use the DNet, the dehazed image remains some haze. When we add CCGAN and CNet respectively, the dehazed images are clearer. When we add CCGAN and CNet simultaneously, the generated images are clearest and they are closest to the corresponding clear images.

\subsection{Qualitative Comparison on Real Hazy Images}

Although the proposed network is trained on synthetic hazy images, we show that it can be generalized to handle real-world hazy images. Fig.~\ref{figure:qualitative-realhaze} shows real hazy images and the corresponding dehazing results generated by state-of-the-art dehazing methods and our method. Although the non-CNN-based dehazing methods are able to remove haze, they excessively remove haze, such as the third row and the fifth column. The CNN-based dehazing methods do not remove haze excessively, but they remain some haze in the images, such as the second row and the sixth, eighth columns. Different from these methods, the images generated by our method shown in the last column are much clearer than those of other methods.

\section{Conclusion} 
\label{sec:conclusion}

In this paper, we propose an unified CNN architecture with the goal to improve the performance both on image dehazing and image classification in an end-to-end learning approach. In comparison to the existing approaches, we investigate the use of class information for synthesizing the dehazed image from a given input hazy image. We evaluate our framework on two benchmark datasets: CUB-200-2011 and Caltech-256. Detailed experiments and comparisons are performed both on synthetic and real-world hazy images to demonstrate that the proposed method significantly outperforms many recent state-of-the-art methods. Additionally, the proposed method is compared against baseline configurations to illustrate the performance gains obtained by introducing the classification sub-network into the framework.


\bibliographystyle{IEEEtran}
\bibliography{ieeetran}


%



\ifCLASSOPTIONcaptionsoff
  \newpage
\fi

\end{document}